\documentclass[journal]{IEEEtran}
\usepackage{cite}

\ifCLASSINFOpdf
   \usepackage[pdftex]{graphicx}
  
\else

\fi

\usepackage{amsmath}

\usepackage{algorithmic}
\usepackage{algorithm}
\usepackage{color}
\makeatletter
\newcommand{\removelatexerror}{\let\@latex@error\@gobble}
\makeatother

\ifCLASSOPTIONcompsoc
  \usepackage[caption=false,font=normalsize,labelfont=sf,textfont=sf]{subfig}
\else
  \usepackage[caption=false,font=footnotesize]{subfig}
\fi

\usepackage{amsfonts}
\usepackage{multirow}
\usepackage{bbding}

\begin{document}
\title{Expansion-Squeeze-Excitation Fusion Network for Elderly Activity Recognition}

\author{Xiangbo~Shu, Jiawen~Yang, Rui~Yan, and Yan~Song
        % <-this % stops a space

 \thanks{X. Shu, J. Yang, Rui Yan, and Y. Song are with the School of Computer Science and
         Engineering, Nanjing University of Science and Technology, Nanjing 210094,
         China. E-mail: \{shuxb, owen, ruiyan, songyan\}@njust.edu.cn. The contribution of the second author equals to the first author. Corresponding Author: R. Yan.}
% of Electrical and Computer Engineering, Georgia Institute of Technology, Atlanta,
% GA, 30332 USA e-mail: (see http://www.michaelshell.org/contact.html).}% <-this % stops a space
% \thanks{J. Doe and J. Doe are with Anonymous University.}% <-this % stops a space
% \thanks{Manuscript received April 19, 2005; revised August 26, 2015.}
}


\markboth{IEEE Transactions on Circuits and Systems for Video Technology, 2022}%
{Shell \MakeLowercase{\textit{et al.}}: Bare Demo of IEEEtran.cls for IEEE Journals}



\maketitle

\begin{abstract}
This work focuses on the task of elderly activity recognition, which is a challenging task due to the existence of individual actions and human-object interactions in elderly activities. Thus, we attempt to effectively aggregate the discriminative information of actions and interactions from both RGB videos and skeleton sequences by attentively fusing multi-modal features. Recently, some nonlinear multi-modal fusion approaches are proposed by utilizing nonlinear attention mechanism that is extended from Squeeze-and-Excitation Networks (SENet). Inspired by this, we propose a novel Expansion-Squeeze-Excitation Fusion Network (ESE-FN) to effectively address the problem of elderly activity recognition, which learns modal and channel-wise Expansion-Squeeze-Excitation (ESE) attentions for attentively fusing the multi-modal features in the modal and channel-wise ways. Specifically, ESE-FN firstly implements the modal-wise fusion with the Modal-wise ESE Attention (M-ESEA) to aggregate discriminative information in modal-wise way, and then implements the channel-wise fusion with the Channel-wise ESE Attention (C-ESEA) to aggregate the multi-channel discriminative information in channel-wise way (referring to Figure~\ref{fig_idea}). Furthermore, we design a new Multi-modal Loss (ML) to keep the consistency between the single-modal features and the fused multi-modal features by adding the penalty of difference between the minimum prediction losses on single modalities and the prediction loss on the fused modality. Finally, we conduct experiments on a largest-scale elderly activity dataset, i.e., ETRI-Activity3D (including 110,000+ videos, and 50+ categories), to demonstrate that the proposed ESE-FN achieves the best accuracy compared with the state-of-the-art methods. In addition, more extensive experimental results show that the proposed ESE-FN is also comparable to the other methods in terms of normal action recognition task. 
\end{abstract}

% Note that keywords are not normally used for peer review papers.

\IEEEpeerreviewmaketitle

\section{Introduction}

\IEEEPARstart{W}{ith} the increasing population of the society, the aging of the population is becoming more and more serious. Coupled with the absence of young people from home all year round, the proportion of empty-nesters in many countries is increasing sharply. Since the physical quality and movement ability of the elderly begin to decline, some kinds of dangers are inclined to occur in life. In such circumstances, some institutions are trying to provide intelligent monitoring for the daily activities of the elderly by employing some advanced technologies in the fields of artificial intelligence and computer vision in recent years. As the key to intelligent monitoring, elderly activity recognition is attracting increased attentions.

By reviewing the normal action recognition methods, we can divide these methods into two categories based on the types of input data, i.e., RGB-based action recognition approach~\cite{shu2020host,li2019collaborative,feichtenhofer2019slowfast,wang2018videos,tang2019coherence,lin2019temporal,carreira2017quo,shu2019hierarchical}, and skeleton-based action recognition approach~\cite{zhang2019view,banerjee2020fuzzy,yan2018spatial,cheng2020skeleton,shu2021spatiotemporal}. For these two types of action recognition methods, various deep neural networks, as the currently main models, have shown remarkable ability to model human actions, such as Convolution Neural Network (CNN)~\cite{albawi2017understanding}, Recurrent Neural Network (RNN)~\cite{zaremba2014recurrent}, Convolutional 3D (C3D)~\cite{tran2015learning}, Long short-term memory (LSTM)~\cite{hochreiter1997long}, Graph Convolutional Networks (GCN)~\cite{kipf2016semi}, and so on. 
Generally, the RGB-based action recognition approach mainly gets motion information of actions from RGB videos, which is disturbed by the background information to some extent. The skeleton-based action recognition approach faces a challenge for recognizing the actions with similar postures. Thus, a natural way is to jointly model motion information from both RGB videos and skeleton sequences~\cite{liu2019action,das2017action}. One impressive solution of these methods is to build two-branch deep neural networks, that firstly learn multi-modal features from RGB and skeleton modalities, and then fuse them. 

\begin{figure}[!t]
		\centering
\includegraphics[scale=0.56]{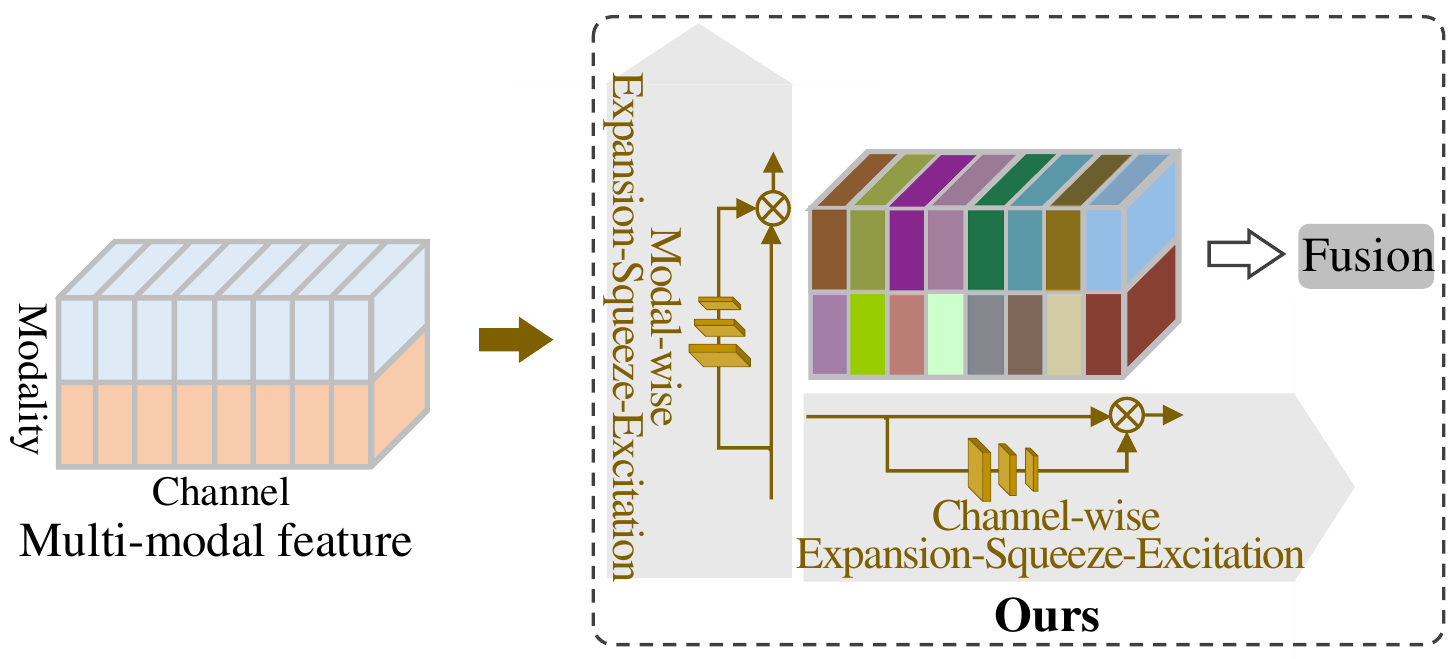}
		\caption{Idea of our method for fusing multi-modal features in this work. Our method fuses multi-modal features with the modal and channel-wise {\bf Expansion-Squeeze-Excitation (ESE)} attentions in the modal and channel-wise ways.}
		\label{fig_idea}
\end{figure}

\begin{figure*}[!t]
		\centering
\includegraphics[scale=0.55]{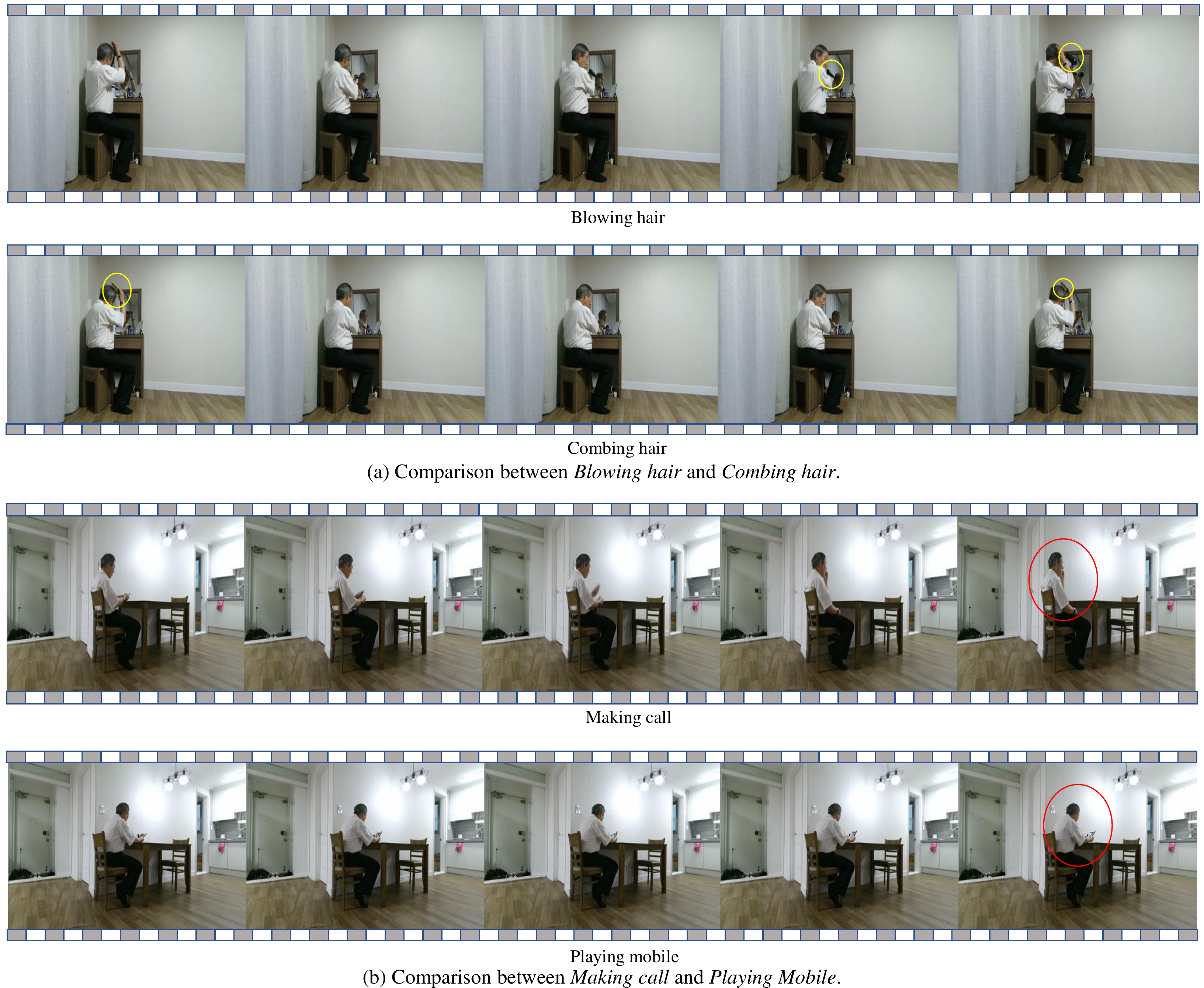}%fig1_new.pdf}
		\caption{Examples of elderly activities. The elderly activities, ``Blowing hair" vs ``Combing hair", as well as ``Making call" vs ``Playing mobile" shows most of similar motions in spatiotemporal space, where only the interacting objects (e.g., comb and blower are marked by yellow circle) are different or only few local movements (right-hand movements are marked by red circle) are distinguishable.}
		\label{fig_elderly_action}
\end{figure*}

Compared with normal action recognition, elderly activity recognition is a more challenging task due to the existence of individual actions and human-object interactions in elderly activities, where many human-object interactions are local; the amplitudes of many elderly activities are unapparent; and the movements of some elderly activities are particularly similar. For example, Figure~\ref{fig_elderly_action} shows two groups of the typical elderly activities, i.e., ``Blowing hair" vs ``Combing hair", and ``Making call" vs ``playing mobile". For the comparison between ``Blowing hair" and ``Combing hair", the motion trajectory and amplitude are very similar in spatiotemporal space, except for the subtle clues of different objects in hand, i.e., comb and blower. For the comparison between ``Making call" and ``Playing mobile", most motions and the interacting object (e.g., mobile) are the same as each other, except for local hand movement in some frames. 
Therefore, how to capture and fuse the discriminative information in RGB and skeleton modalities is crucial for modeling elderly activities.

 %模型图
\begin{figure*}[t]
\centering
\includegraphics[scale=0.55]{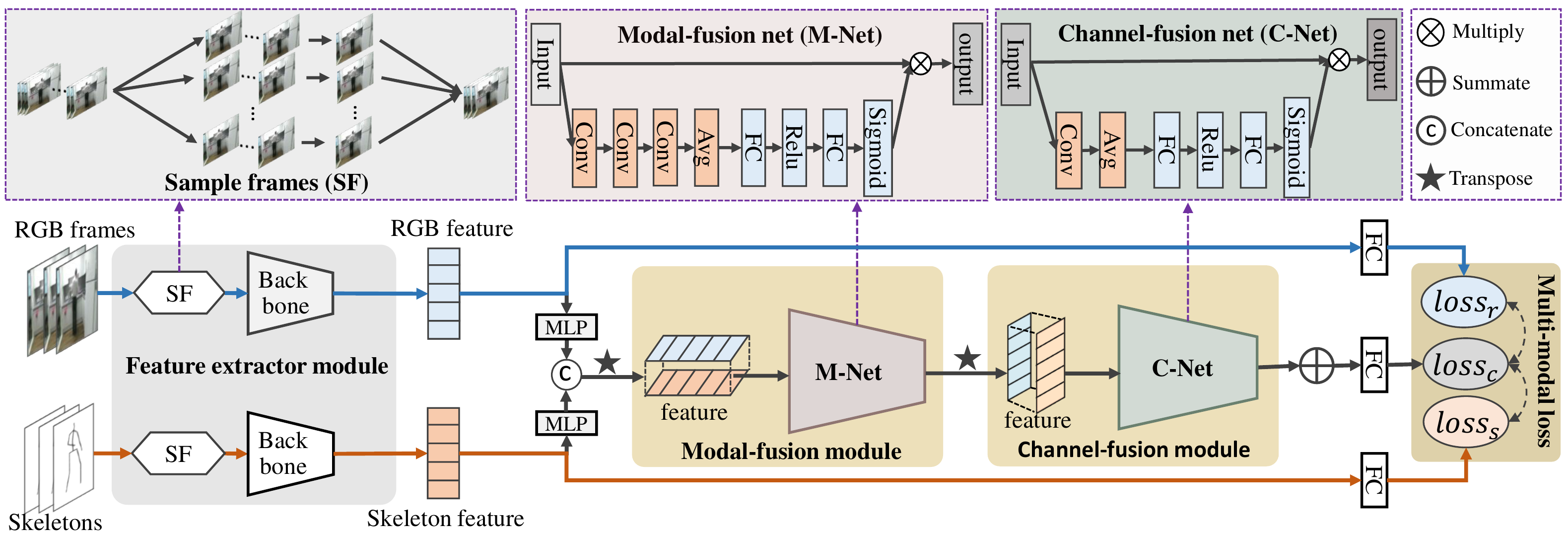}
\caption{The framework of the proposed ESE-FN method mainly consists of feature extractor module, modal-fusion module, channel-fusion module, and multi-modal loss. It firstly employs the backbone to extract the RGB and skeleton features, respectively. And then, the concatenated RGB and skeleton features are fed into the modal-fusion module and channel-fusion module for the modal and channel-wise fusions in turn. Specifically, M-Net and C-Net aim to fuse features with the modal and channel-wise expansion-squeeze-excitation attentions for capturing modal and channel-wise dependencies among features, respectively. Finally, a new multi-modal loss containing three sub-losses is designed to optimize the whole network by measuring the consistency between the prediction loss on fused modalities and the prediction loss on single modalities.}
\label{fig_framework}
\end{figure*}

Multi-modal data or features fusion across different modalities has always been a hot topic for years~\cite{nojavanasghari2016deep,wang2020deep,wang2021multi,tang2018multi,joze2020mmtm,zhu2019multi,gao2021unified,zhang2020multi,zhou2021ecffnet,zhang2020revisiting}. Since multi-modal features often contain irrelevant (modal-specific) information, the straightforward feature-level or score-level fusion will degrade the discriminative ability of the fused features. The key to successful fusion is how to reinforce the discriminative information while suppressing the irrelevant information among multi-modal features. To this end, some works~\cite{francis2019fusion,hori2017attention,liu2018efficient} propose to utilize the linear attention to selectively fuse multi-modal features. Witnessing the impressive classification performance of Squeeze-and-Excitation Networks (SENet)~\cite{hu2018squeeze} on ImageNet, some researchers~\cite{he2018exploiting,kuang2019multi,fooladgar2019multi,liu2020attentive} extend linear fusion to the nonlinear fusion via Squeeze-and-Excitation (SE).

In this work, we consider designing a couple of Squeeze-and-Excitation based networks as nonlinear attention mechanisms for effectively aggregating the motion information in video and skeleton modalities by additionally bringing in expansion, called Expansion-Squeeze-Excitation (ESE). Based on this, we propose a novel Expansion-Squeeze-Excitation Fusion Network (ESE-FN) to fuse multi-modal features with modal and channel-wise ESE attentions. The overview framework of ESE-FN is shown in Figure~\ref{fig_framework}, which mainly consists of four parts, i.e., feature extractor module, modal-fusion module, channel-fusion module, and multi-modal loss. First, ESE-FN randomly samples RGB frames in different video clips, which are fed into the backbone to extract the RGB features. Likewise, the skeleton features can be also extracted from sampled skeleton sequences. Second, the concatenated RGB and skeleton features are fed into the modal-fusion module for attentively fusing features in modal-wise way, where M-Net aims to learn the Modal-wise Expansion-Squeeze-Excitation Attention (M-ESEA) via modal-wise ESE, as shown in Figure~\ref{fig_idea}. Third, the output from the modal-fusion module is further fed into the channel-fusion module for attentively fusing features in channel-wise way, where C-Net aims to learn the Channel-wise Expansion-Squeeze-Excitation Attention (C-ESEA) via channel-wise ESE. Finally, we utilize three types of features, i.e., RGB features, skeleton features, fused multi-modal features, to construct three sub-losses, which are integrated into a new Multi-modal Loss (ML). Here, ML additionally brings in the penalty of difference between the minimum prediction losses on single modalities and the prediction loss on the fused modality, which can keep the consistency between the single-modal features and the fused multi-modal features.

	Overall, the main contributions of this work can be summarized as follows.
	\begin{itemize}
		\item We deeply explore the characteristics of elderly activities, and propose a novel Expansion-Squeeze-Excitation Fusion Network (ESE-FN) to attentively fuse multi-modal features in the modal and channel-wise ways for effectively addressing the problem of elderly activity recognition. 
		\item To well capture the discriminative information of multi-modal features, we design a flexible Modal-fusion Net (M-Net) and Channel-fusion Net (C-Net) to learn Modal-wise Expansion-Squeeze-Excitation Attention (M-ESEA) and Channel-wise Expansion-Squeeze-Excitation Attention (C-ESEA) for capturing the modal and channel-wise dependencies among features, respectively.
		\item To keep the consistency between the single-modal features and the fused multi-modal features, we design a new Multi-modal Loss (ML) to additionally measure the difference between the minimum prediction loss on single modalities and the prediction loss on the fused modality.
		\item By conducting extensive experiments on both elderly activity recognition and normal action recognition tasks, we illustrate that the proposed ESE-FN method achieves the SOTA performance compared with the other competitive methods.
	\end{itemize}
	
\par
The rest of this paper is organized in the following. Section~\ref{RW} surveys some works related to RGB-based action recognition, skeleton-based action recognition, and multi-modal fusion. Section~\ref{M} introduces the proposed method for elderly activity recognition in details. Section~\ref{E} presents results and analysis of experiments, followed by the conclusions in Section~\ref{C}. 

\section{Related Work}
We survey some works related to RGB-based action recognition, skeleton-based action recognition, and multi-modal fusion.
\label{RW}
\subsection{RGB-based Action Recognition}
Most existing RGB-based Action Recognition methods~\cite{shu2016image,ji2019image,shu2017concurrence,zhang2020data,tang2019social,yan2018participation,shu2020host,yan2020higcin,song2020bi,jiang2020cross,jin2015partially,qi2018dotanet} can be categorized into two classes. The first class is based on a two-stream network that usually uses RGB and optical flow to model spatial and temporal information, respectively. Karen et al.~\cite{simonyan2014two} proposed a two-stream network to model spatial and temporal information in RGB and optical flow frames for the first time. Subsequently, Wang et al.~\cite{wang2016temporal} proposed a Temporal Segment Network (TSN) to model long-range temporal action. Crasto et al.~\cite{crasto2019mars} proposed a distillation network that uses optical flow to distill RGB data, which can make the RGB-based model learn the temporal information. It is well known that optical flow is computed by using the RGB frames, which is time-consuming and would bring in a bottleneck. The second class is based on a series of 3D convolutional networks, such as, C3D~\cite{tran2015learning}, I3D~\cite{carreira2017quo}, T3D~\cite{diba2017temporal}, Res3D~\cite{tran2017convnet}, and so on, which are extended from 2D networks in spatiotemporal dimension. Due to the computation consumption of general 3D convolutional networks, Qiu et al.~\cite{qiu2017learning} proposed a Pseudo-3D residual network (P3D) that decomposes the convolutions into separate 2D spatial and 1D temporal filters. Moreover, Feichtenhofer et al.~\cite{feichtenhofer2019slowfast} proposed to use two different frame rates to accelerate train 3D networks.

\par
The above-mentioned methods are proposed to address the problem of normal action recognition. Compared with the normal action recognition task, the elderly activity recognition task requires more discriminative information to identify some subtle individual actions and human-object interactions in elderly activities. Due to the lack of temporal information on RGB data, the network implementing on RGB data cannot accurately describe the actions with obvious temporal information. Although some temporal information can be captured from optical flow data, the calculation of optical flow is too time-consuming.%Thus, more and more researchers pay attention to the Skeleton-based Action Recognition.

\subsection{Skeleton-based Action Recognition}
So far, there are many action recognition works based on skeleton~\cite{shu2021spatiotemporal,zhang2019view,banerjee2020fuzzy,yan2018spatial,banerjee2020fuzzy,shao2020learning,song2017multi}, called skeleton-based action recognition methods. The main target of skeleton-based action recognition is to learn temporal and spatial information from skeleton sequences. In the early stages, researchers utilized various deep neural networks. e.g., Recurrent Neural Network (RNN), Convolution Neural Network (CNN), and Long Short-Term Memory (LSTM), to model skeleton motions for capturing temporal and spatial information. For example, Zhang et al.~\cite{zhang2019view} proposed an adaptive model designed by CNN and RNN to solve the problem of view difference during the skeleton data collection and action shooting. Banerjee et al.~\cite{banerjee2020fuzzy} used CNN to extract different complementary motion information from skeleton sequences, e.g., distance and angle between joint points, to better model the temporal information of skeleton data. Recently, researchers used Graph Convolution Network (GCN) to model temporal and spatial information from skeleton sequences by treating skeletons, joints, and bonds as graphs, nodes, and edges, respectively~\cite{yan2018spatial,liu2020multi,cheng2020skeleton}. For example, Yan et al.~\cite{yan2018spatial} proposed a Spatial-Temporal Graph Convolutional Network (ST-GAN) that employs graph convolution to aggregate the joint features in the spatial dimension. Liu et al.~\cite{liu2020multi} proposed a multi-stream graph convolutional network to avoid the missing of structural information in the training phase. Cheng et al.~\cite{cheng2020skeleton} proposed a Shift Graph Convolutional Network (Shift-GCN) for solving the problem of inflexible acceptance domain of graph convolution network on both spatial and temporal dimensions.

It can be found that skeleton-based action recognition methods mainly design an effective model to capture temporal and spatial information from skeleton sequences, which is difficult to recognize human-object actions, e.g., blowing hair, combing hair, etc. %Since some elderly movements have subtle differences in skeleton sequences, it will be extremely difficult to identify the actions of the elderly only from skeleton data. 

\subsection{Multi-modal Fusion}
Many works~\cite{liu2021learning,chen2019novel,liu2019action,das2017action,chen2016fusion,zhu2019multi,wang2021multi,tang2018multi,zhang2020revisiting,zhou2021ecffnet,gu2018two,jiang2020cross,chen2021attention,song2020bi,jin2018deep,shi2020can,zhao2022wavelet} consider aggregating information from multi-modal data for various image and video content analysis tasks, such as video action recognition, video anomaly detection and localization, and so on. For example, Wang et al.~\cite{wang2021multi}  and Tang et al.~\cite{tang2018multi} proposed the score-level and feature-level multi-modal fusions, respectively. Those straightforward fusion methods did not consider the interaction among different multi-modal information. Zhang et al.~\cite{zhang2020revisiting} proposed to employ several convolutions to fuse the RGB and depth features, which only considered the interaction among multi-modal information from the local view. Zhou et al.~\cite{zhou2021ecffnet} proposed to fuse the RGB and thermal image features by channel and spatial-wise attentions, respectively. Overall, on the one hand, different modality data have different distributions. On the other hand, multi-modal data often contain irrelevant (modal-specific) information. Thus, how to aggregate information from multi-modal features is the main problem. Multi-modal data or features fusion across different modalities is always a hot topic, some classical methods~\cite{francis2019fusion,hori2017attention,liu2018efficient} were proposed to utilize the linear embedding or attention mechanism to fuse multi-modal features. For example, Hori et al.~\cite{hori2017attention} propose a multi-modal attention model that selectively fuses multi-modal features based on learned attention. With the advancement of Convolutional Neural Networks (CNN), some nonlinear fusion approaches of multi-modal features are proposed by designing various convolutional networks as the attention mechanism~\cite{he2018exploiting,kuang2019multi,fooladgar2019multi,liu2020attentive,raza2020pfaf}. For example, Kuang et al.~\cite{kuang2019multi} proposed a multi-modal fusion network based on CNN for face anti-spoofing detection. Fooladgar et al.~\cite{fooladgar2019multi} used an efficient attention method based on CNN architecture to fuse RGB data and depth maps in channel-wise way. Su et al.~\cite{su2020msaf} used the soft attention method to interact multi-modal data in channel-wise way. To fully fuse the multi-modal features, we consider interacting information of multi-modal features not only in channel-wise way but also in modal-wise way.  In this work, we design a flexible Expansion-Squeeze-Excitation (ESE) in Modal-fusion Net (M-Net) and Channel-fusion Net (C-Net) to learn modal and channel-wise nonlinear attention for capturing the modal and channel-wise dependencies among features, respectively.
% {\color{blue}
% He et al.~\cite{he2018exploiting} propose an improved temporal Xception network to integrate multi-modal information by combining both the early and later fusion of multiple modalities.
% In this work, we design a flexible Expansion-Squeeze-Excitation (ESE) in Modal-fusion Net (M-Net) and Channel-fusion Net (C-Net) to learn modal and channel-wise nonlinear attention for capturing the modal and channel-wise dependencies among features, respectively.
% }

\section{Methodology}
\label{M}
In this section, we mainly introduce Expansion-Squeeze-Excitation Fusion Network (ESE-FN). Specifically, we first revisit the SENet as a warm-up, and then introduce the framework of ESE-FN in details, including Modal-fusion Net (M-Net), Channel-fusion Net (C-Net), and Multi-modal Loss (ML).

\subsection{Revisiting SENet}
\label{RS}
Squeeze-and-Excitation Networks (SENet)~\cite{hu2018squeeze} has shown remarkable performance in the ImageNet database by explicitly capturing the channel-wise dependencies between feature maps. Here, the channel-wise dependencies are quantified via the nonlinear attention corresponding to each channel. The nonlinear attention is obtained by Squeeze-and-Excitation (SE). SENet has been proven that learning channel-wise nonlinear attention can improve the discriminative ability of features. 

In SENet, Squeeze-and-Excitation (SE) can be divided into two steps: squeeze and excitation in turn, which are used for obtaining channel-wise representation and channel-wise nonlinear attention. Specifically, assuming $C$ feature maps, denoted by $X=\{X_c\}_{c=1}^{C}$, for one feature map $X_c$ in the $c$-th channel, the squeeze operation is performed as follows,
\begin{equation}
Y_c=\frac{1}{H\times W}\sum_{i=1}^{H}\sum_{j=1}^{W} X_c(i,j),
\end{equation}
where $H$ and $W$ denote the height and width of the feature map, respectively, $Y_c$ is the result of global average pooling for the $c$-th channel.  
In the squeeze step, all feature maps are transformed to a channel-wise representation $Y=[Y_1,Y_2,\cdots,Y_c,\cdots,Y_C]^T$, which is a one-dimensional vector. In the excitation step, the channel-wise representation $Y$ is fed into multi fully-connected layers to obtain the channel-wise attention $W$, as follows,
\begin{equation}
W=\sigma(F_1({\text {Relu}}(F_2(Y)))),
\end{equation}
where ${F_*}$ denotes a Fully Connected (FC) layer in this work, and ${\sigma}$ is a sigmoid function. Finally, the original feature maps $X$ can be updated to $\hat {X}$ by the following equation
\begin{equation}
\hat{X}=X \otimes W,
\end{equation}
where $\otimes$ denotes channel-wise multiplication between $X$ and $W$. Compared with the original feature maps $X$, the updated feature map $\hat{X}$ has enhanced the discriminative information, while suppressing the useless or irrelevant modal-specific information to some extent. 
\par

Through the above analysis, we can find that SENet uses the global average pooling to interact spatial information. Since the global average pooling is implemented from the global view, the learned nonlinear attention is significantly affected by noise (e.g., a large area of background). In this work, to well capture the local and global spatial information of features, we consider interacting spatial information from the local and global views.

\subsection{Overview of ESE-FN}
For the task of elderly activity recognition, we attempt to aggregate the discriminative information from both RGB videos and skeleton sequences by attentively fusing multi-modal features. Thus, multi-modal features fusion across multiple modalities is key to modeling the motion among RGB videos and skeleton sequences. Based on Squeeze-and-Excitation in SENet, we design
a new Modal-fusion Net (M-Net) and a new Channel-fusion Net (C-Net) to learn the Modal-wise Expansion-Squeeze-Excitation Attention (M-ESEA) and Channel-wise Expansion-Squeeze-Excitation Attention (C-ESEA) for capturing the modal and channel-wise dependencies among features, respectively. By integrating M-Net and C-Net as the main components, a novel Expansion-Squeeze-Excitation Fusion Network (ESE-FN) is proposed to attentively fuse the multi-modal features with M-ESEA and C-ESEA. The framework of ESE-FN is shown in Figure \ref{fig_framework}, which mainly consists of four parts, i.e., feature extractor module, modal-fusion module, channel-fusion module, and multi-modal loss.

Denoted ${\left \{ v_i|i=1,2,3,... N \right \}}$ as an video set with size $N$, we first split each video into ${T}$ clips and random sample a frame from each clip. Then, we get the frame set ${\left \{ r_i|i=1,2,3,... T \right \}}$ as one input video ${r}$, which are fed into the RGB backbone (e.g., ResNeXt101~\cite{xie2017aggregated}) to extract the RGB feature ${f_r \in {R}^{d_1\times 1}}$ ( $d_*$ denotes the feature dimension in this paper), as follows,

\begin{equation}
f_r={F_\text{RGB}}(r).
\end{equation}

Likewise, for video $r$, the corresponding skeleton sequence $s$ is obtained and fed into the skeleton backbone (e.g., Shif-GCN~\cite{cheng2020skeleton}) to extract the skeleton feature ${f_s \in {R}^{d_2 \times 1}}$,

\begin{equation}
f_s={F_\text{Skeleton}}(s).
\end{equation}

Generally, the multi-modal features, i.e., $f_r$ and $f_s$, have different sizes and cannot be directly concatenated. Thus, we use two MLPS to unify the size of $f_r$ and $f_s$, and then concatenate them, as follows,

\begin{equation}
f={F_\text{Concat}}(H_r(f_r),H_s(f_s)),
\end{equation}

where ${f \in {R}^{d \times n}}$, $n$ is the number of modality, ${H}_r$ and $H_s$ are MLP, and ${{\text {Concat}}(\cdot)}$ is a modal-wise concatenation operator. 

Subsequently, we transpose the feature $f$ and feed it into M-Net for modal-wise fusion, as follows,

\begin{equation}
h_{m}={F_\text{M-Net}}({F_\text{Trans}}(f)),
\end{equation}

where ${F_\text{Trans}{(\cdot)}}$ is a transpose operator.
Likewise, $h_{m}$ is transposed and fed into C-Net for channel-wise fusion, as follows,

\begin{equation}
h_{mc}={F_\text{C-Net}}({F_\text{Trans}}(h_m)),
\end{equation}

We sum up feature $h_{mc}$ in modal-wise way and get the fused multi-modal feature $f_{rs}$.

\begin{figure*}[ht]
\centering
\includegraphics[scale=0.56]{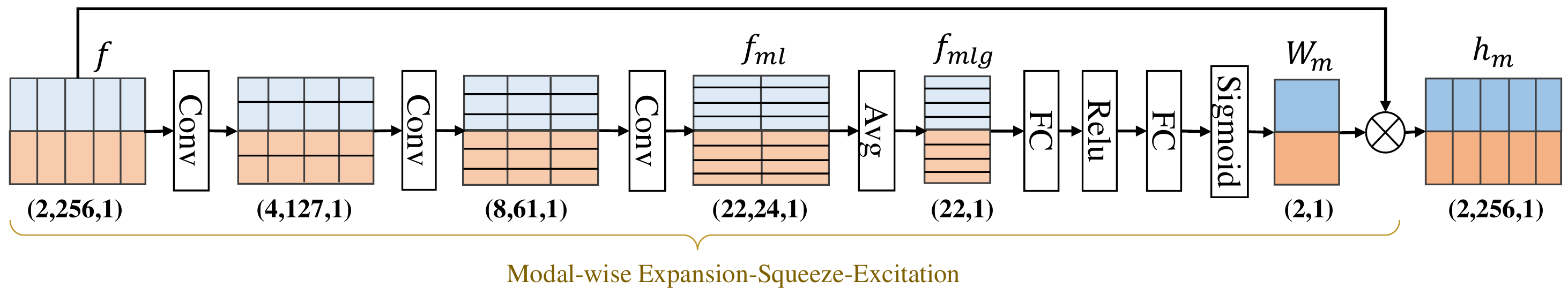}
\caption{Configuration of Modal-fusion Net (M-Net) in details. The multi-modal features are fused with Modal-wise Expansion-Squeeze-Excitation Attention (M-ESEA)
in modal-wise way. M-ESEA is obtained by the modal-wise nonlinear attention mechanism mainly composed of the convolutional layers, average pooling layer, and FC layer.}
\label{fig_mnet}
\end{figure*}

Finally, RGB feature $f_r$, skeleton feature $f_s$ and fused multi-modal feature $f_{rs}$ are fed into a new Multi-modal Loss (ML) for optimizing all parameters of ESE-FN, as follows,
\begin{equation}
\Theta={\text {argmin}}({\mathcal {L}}(F_r(f_r),F_{rs}(f_{rs}),F_s(f_s))),
\end{equation}
where $\Theta$ denotes the parameter set of the proposed ESE-FN. The following sections introduce M-Net, C-Net, Multi-modal Loss in details.

\subsection{Modal-fusion Net (M-Net)}
 The detailed configuration of Modal-fusion Net (M-Net) is shown in Figure \ref{fig_mnet}. M-Net can attentively aggregate the local and global spatial discriminative information of multi-modal features via M-ESEA in modal-wise way. Specifically, we firstly extend the transposed feature ${f\in {R}^{n \times d}}$ to ${f\in {R}^{n \times d \times 1}}$ as the input of M-Net. Similar to Squeeze-and-Excitation in SENet, we divide the implementation of M-Net into three steps, i.e., modal-wise expansion, modal-wise squeeze, and modal-wise excitation in turn, which are detailed in the following.
%Similar to SENet, we also divide the implement of our M-Net into two steps, i.e., modal-wise squeeze and modal-wise excitation in turn, which are detailed in the following.

{\bf Modal-wise expansion step.} %It uses several stacked convolutional layers with different kernels to interact modal-wise information by expanding the spatial information from the local view.
It uses several stacked convolutional layers with different kernels to interact modal-wise information by expanding the modal information from the local view.
Then the expanded feature ${f_{ml}}$ can be described as follows,
%{\bf Modal-wise squeeze step.} First, several convolutional layers with different kernels (as shown in Figure \ref{fig_mnet}) interacts spatial information from the local view. Then the interacted features ${f_{ml}}$ can be described as follows,
\begin{equation}
f_{ml}={\text {Conv}}_1({\text {Conv}}_2({\text {Conv}}_3(f))),
\end{equation}
%where ${f_{ml} \in {R}^{m \times d_m \times 1}}$, ${m > n}$, $d_m$ is the spatial dimension after convolution transformation (${d_m < d}$). Here, ${f_{ml}}$ can be also regarded as the single-modal feature set $\{f_{ml}^{i} \in R^{(m/n) \times d_m}\}_{i=1}^{n}$. Then we can use ${f_{ml}}$ to calculate the modal-wise representation ${{f_{mlg}} \in {R}^{m\times 1}}$ from global view via the average pooling, as follows,
where ${f_{ml} \in {R}^{m \times d_m \times 1}}$, $d_m$ is the spatial dimension after convolution transformation (${m \times d_m > n\times d}$). Here, ${f_{ml}}$ can be also regarded as the single-modal feature set $\{f_{ml}^{i} \in R^{(m/n) \times d_m}\}_{i=1}^{n}$. 

{\bf Modal-wise squeeze step.} It utilizes ${f_{ml}}$ to calculate the modal-wise representation ${{f_{mlg}} \in {R}^{m\times 1}}$ (${m < n\times d}$) from global view via the average pooling, as follows,
\begin{equation}
{f_{mlg}}=\frac{1}{d_m \times 1} \sum_{i=1}^{d_m} \sum_{j=1}^{1} f_{ml}(i,j).
\end{equation}

{\bf Modal-wise excitation step.} It utilizes the modal-wise representation ${f_{mlg}}$ to learn the Modal-wise Expansion-Squeeze-Excitation Attention (M-ESEA) ${W_m}$ for capturing the modal-wise dependence of features. Finally, the original input feature $f$ can be updated to the modal-wise fused feature ${h_m \in {R}^{n\times d\times 1}}$ based on the following equations,
\begin{equation}
W_m=\sigma(F_3({\text {Relu}}(F_4({f_{mlg}}))));
\end{equation}
\begin{equation}
h_m= f\otimes W_m.
\end{equation}

Compared with SENet, it is noted that we additionally bring in the expanding of features as the partner of the squeezing. Expanding and squeezing provide the interactions of features in up-size and down-size ways, which can capture both of local and global dependencies of features.
Different from SENet dealing with the single-modal feature map, we deal with the multi-modal features which can been seen as a feature map by concatenating two modal feature vectors. In the modal-wise expansion step, the convolutions with different kernels can be used to capture the local spatial correlation of multi-modal features in different regions, which can capture the local dependencies of features. In the modal-wise squeeze step, the global average pooling can be used to capture the global spatial correlation of multi-modal features, which can capture the global dependencies of features.

\begin{figure}[!t]
\centering
\includegraphics[scale=0.41]{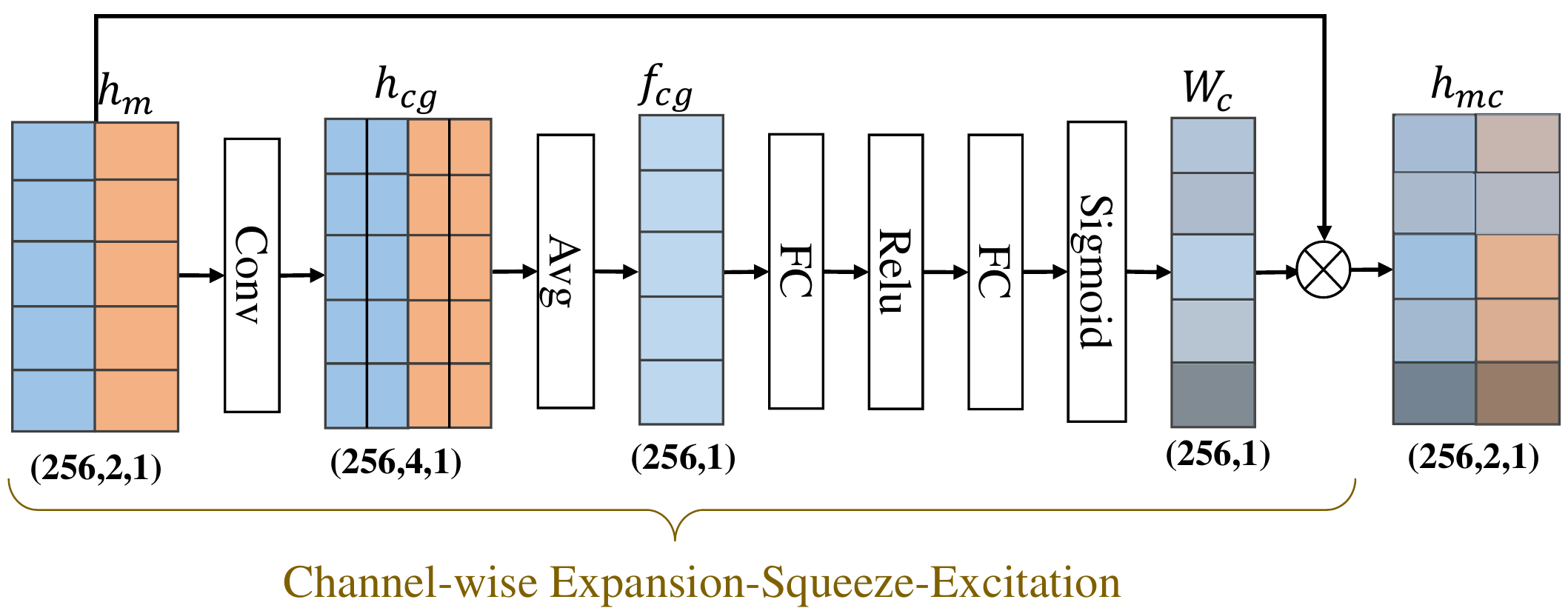}
\caption{Configuration of Channel-fusion Net (C-Net) in details. The multi-modal features are fused with Channel-wise Expansion-Squeeze-Excitation Attention (C-ESEA) in channel-wise way. C-ESEA is obtained by the channel-wise nonlinear attention mechanism mainly composed of the convolutional layer, average pooling layer, and FC layer.}
\label{fig_cnet}
\end{figure}

\subsection{Channel-fusion Net (C-Net)}
In this section, we focus on our Channel-fusion Net (C-Net), as shown in figure \ref{fig_cnet}. C-Net aims to learn C-ESEA for fusing multi-modal features in channel-wise way. Similar to M-Net, the implementation of C-Net can be also divided into the channel-wise expansion, channel-wise squeeze, and channel-wise excitation in turn, described in the following. %the channel-wise squeeze and channel-wise excitation steps. 

{\bf Channel-wise expansion step.} %It uses the convolutional layer to interact the channel-wise information by expanding spatial information from the local view. 
It uses the convolutional layer to interact channel-wise information by expanding the channel information from the local view.
Then the expanded features ${h_{cg}}$ can be calculated as follows,
\begin{equation}
h_{cg}=\text{Conv}_4(h_m),
\end{equation}
where ${h_{cg} \in R^{d\times n_1\times 1}}$, ${n_1 > n}$. It is noted that the multi-modal features are processed by M-Net and C-Net in turn. First, in M-Net, the difference of cross-modal features is relatively large, thus the expansion is large by setting multi-layers. Subsequently, in C-Net (after processed by M-Net), the difference of cross-modal features is relatively small, which only requires a relatively smaller expansion. Thus we set only one convolution layer in C-Net for lightweight designing.

{\bf Channel-wise squeeze step.} It utilizes ${h_{cg}}$ to calculate the channel-wise representation ${f_{cg}}$ from the global view via the average pooling, as follows,
\begin{equation}
f_{cg}=\frac{1}{n_1 \times 1} \sum_{i=1}^{n_1} \sum_{j=1}^{1} h_{cg}(i,j),
\end{equation}
where, ${f_{cg}}\in R^{d\times 1}$ can be also regarded as the single-channel feature set $\{f_{cg}^i\in R^{1\times 1} \}_{i=1}^{d}$.
%where ${h_{cg} \in R^{d\times n_1\times 1}}$, ${n_1 > n}$. Here, ${f_{cg}}$ can be also regarded as the single-channel feature set $\{f_{cg}^i\in R^{1\times 1} \}_{i=1}^{d}$. 

{\bf Channel-wise excitation step.} It utilizes the channel-wise representation ${f_{cg}}$ to learn the Channel-wise Expansion-Squeeze-Excitation Attention (C-ESEA) ${W_c}$ for capturing the channel-wise dependence of features, as follows, 
\begin{equation}
W_{c}=\sigma(F_5({\text{Relu}}(F_6(f_{cg})))).
\end{equation}
Finally, we get the feature updated in channel-wise way, described as follows,
\begin{equation}
h_{mc}=h_{m} \otimes W_{c}.
\end{equation}

By comparing with the formulations of M-Net and C-Net, modal-wise fusion can be seen as a rough fusion of multi-modal features in modal-wise way, while channel-wise fusion can be seen as a fine-grained fusion of multi-modal features in channel-wise way. Both of them attentively aggregate the discriminative information of multi-modal features based on modal and channel-wise dependencies of features.

\subsection{Multi-modal Loss (ML)}

To keep the consistency between the single-modal features and the fused multi-modal features, we design a new Multi-modal Loss (ML) to additionally measure the difference between the prediction loss on single modalities and the prediction loss on the fused modality. The key idea of multi-modal loss is that we take the minimum prediction loss on single modalities to be consistent with the prediction loss on the fused modality. Formally, we define three types of recognition losses ${\mathcal L}_{r}$, ${\mathcal L}_{s}$, and ${\mathcal L}_{rs}$ corresponding to the RGB modality, skeleton modality, and fused modality, respectively. Then the multi-modal loss can be described as follows, 
\begin{equation}
{\mathcal L}=\alpha \times {\mathcal L}_{rs}+\beta \times ({\text {min}}({\mathcal L}_{r},{\mathcal L}_{s})-{\mathcal L}_{rs}),
\label{eq18}
\end{equation}
where ${\alpha}$ and ${\beta}$ are hyper-parameters (will be discussed in Section~\ref{DS}). In this work, the forms of ${\mathcal L}_{r}$, ${\mathcal L}_{s}$, and ${\mathcal L}_{rs}$ are cross entropy loss.

\section{Experiments}
\label{E}
 In this section, we conduct experiments to evaluate the performance of the proposed ESE-FN in terms of the elderly activity recognition task. Besides, we further conduct comparative experiments between the proposed ESE-FN and the other advanced methods in terms of the normal action recognition task. 

\subsection{Datasets}
\label{train-detail}
We evaluate the performance of ESE-FN in terms of the elderly activity recognition task on the ETRI-Activity3D dataset~\cite{jang2020etri}. It is the currently largest elderly activity recognition dataset collected in real-world surveillance environments, which contains 112,620 samples performed by 100 persons including RGB videos, depth maps, and skeleton sequences. All videos are grouped into 55 classes of actions, including individual activities, human-object interactions, and multiperson interactions. The splitting of training and testing sets is based on person ID, namely the samples with person ID $\{3, 6, 9, , \cdots, 99\}$ for testing, and the samples with person ID $\{1,2,4,5, \cdots, 100\}$ for training. 

\subsection{Implementation details}
In the data pre-processing phase, we split each video into $T=64$ clips, and randomly select one frame for each clip. Finally, we obtain a new video set, wherein each video contains $64$ frames. Likewise, we also obtain a new skeleton set, wherein each skeleton sequence contains $64$ frames. 

In the feature extraction phase, we use ResNeXt18 or ResNeXt101~\cite{xie2017aggregated} as the RGB backbone, and Shift-GCN~\cite{cheng2020skeleton} as the skeleton backbone. 
For the pre-trained ResNeXt18 and ResNeXt101, we fine tune them with the standard SGD optimizer by setting the momentum, initial learning rate, weight decay, and total epochs as  0.9, 0.1, ${10^{-3}}$, and 120, respectively. Especially, the batch size is set to 128 and 32 for ResNeXt18 and ResNeXt101, respectively.  
For the pre-trained Shift-GCN, we fine tune it with the standard SGD optimizer by setting the momentum, initial learning rate, batch size, and total epochs as 0.9, 0.1, ${10^{-4}}$, 32, and 140, respectively. Subsequently, we use the fine-tuned RGB and skeleton backbone to extract RGB and skeleton features, respectively. Here, to test the representation performance of ResNeXt18, ResNeXt101, and Shift-GCN, we use these three models to extract single-modal features for training a softmax independently. For example, we use ResNeXt18 to extract RGB features and feed them into the softmax. The obtained performance for three models is listed in Table~\ref{tab_backbone}. We can see that ResNeXt101 performs better than ResNeXt18. In this paper, we choose ResNeXt101 as the RGB backbone in default.

\begin{table}[!t]
\centering
\caption{{Test on the representation power for different backbones on the ETRI-Activity3D dataset.}}
\label{tab_backbone}
\begin{tabular}{l|c|c|c}
\hline
\hline
Modality              & Backbone         & Params         & Accuracy (\%)   \\ \hline
\multirow{2}{*}{RGB}  & ResNeXt18        & 15.60M         & 87.1          \\
                      & ResNeXt101       & 48.16M         & 93.5            \\ \hline
Skeleton              & Shift-GCN        & 0.42M           & 88.6            \\ \hline \hline
\end{tabular}
\end{table}

\begin{figure}[!t]
\centering
\includegraphics[scale=0.45]{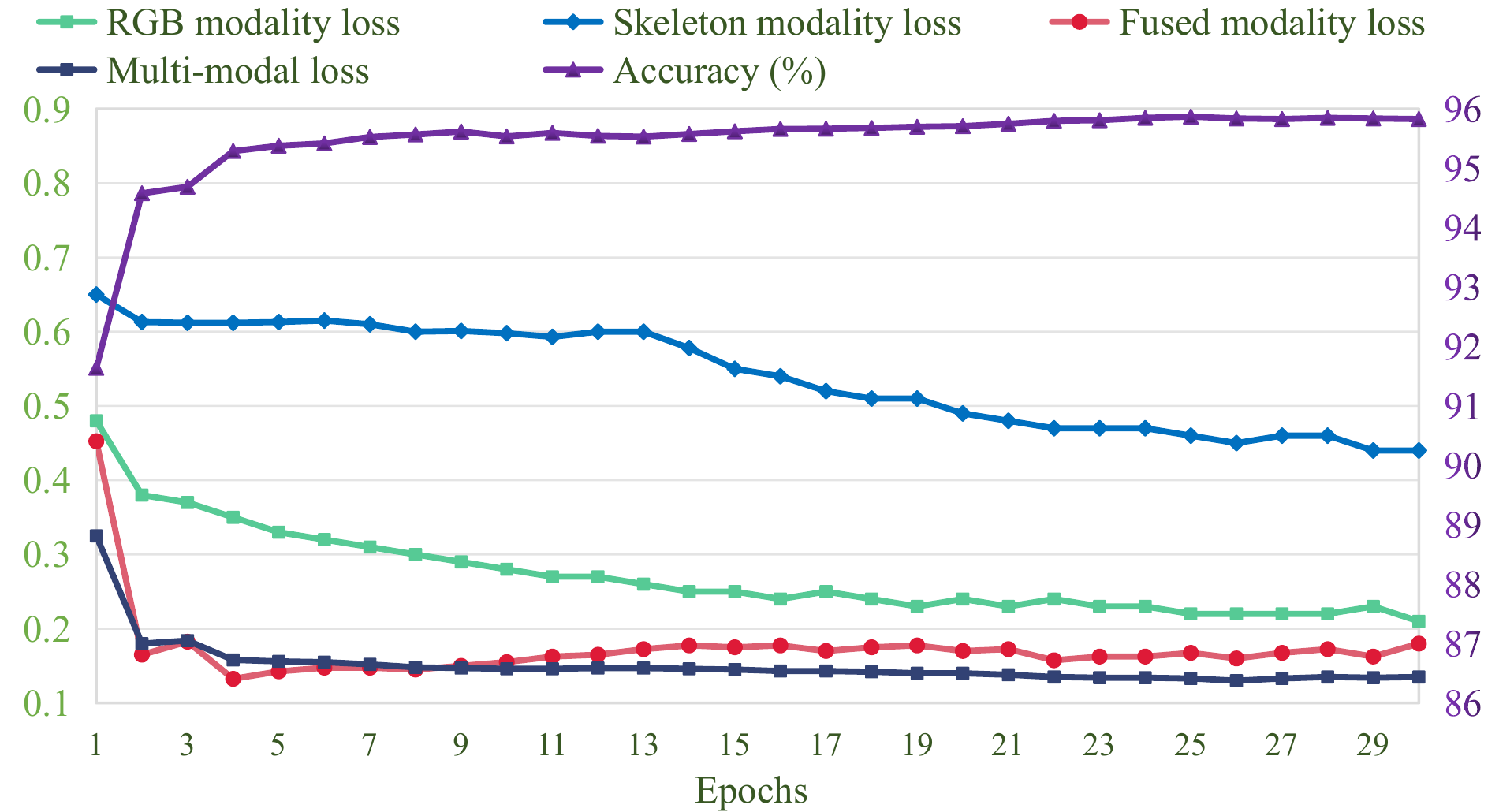}
\caption{{The training loss and accuracy of ESE-FN on the ETRI-Activity3D dataset.}}
\label{fig_loss_acc}
\end{figure}

In the training phase of ESE-FN, we use the standard SGD optimizer to train it by setting the momentum, basic learning rate, weight decay, batch size, and total epochs as 0.9, 0.1, ${10^{-4}}$, 32 and 30, respectively. All above experiments are performed via the PyTorch deep learning framework on the Linux server equipped with Titan RTX GPU. The training loss and accuracy of ESE-FN as the number of epochs are shown in Figure~\ref{fig_loss_acc}. We can see that the training loss and accuracy of ESE-FN reach a steady state after about $30$ epochs. And then, the overall loss is consistent with the other sub-losses (i.e., RGB modality, skeleton modality, and fused modality losses) in terms of convergence.

\subsection{Diagnostic Study}
\label{DS}
\begin{figure}[!t]
\centering
\includegraphics[scale=0.40]{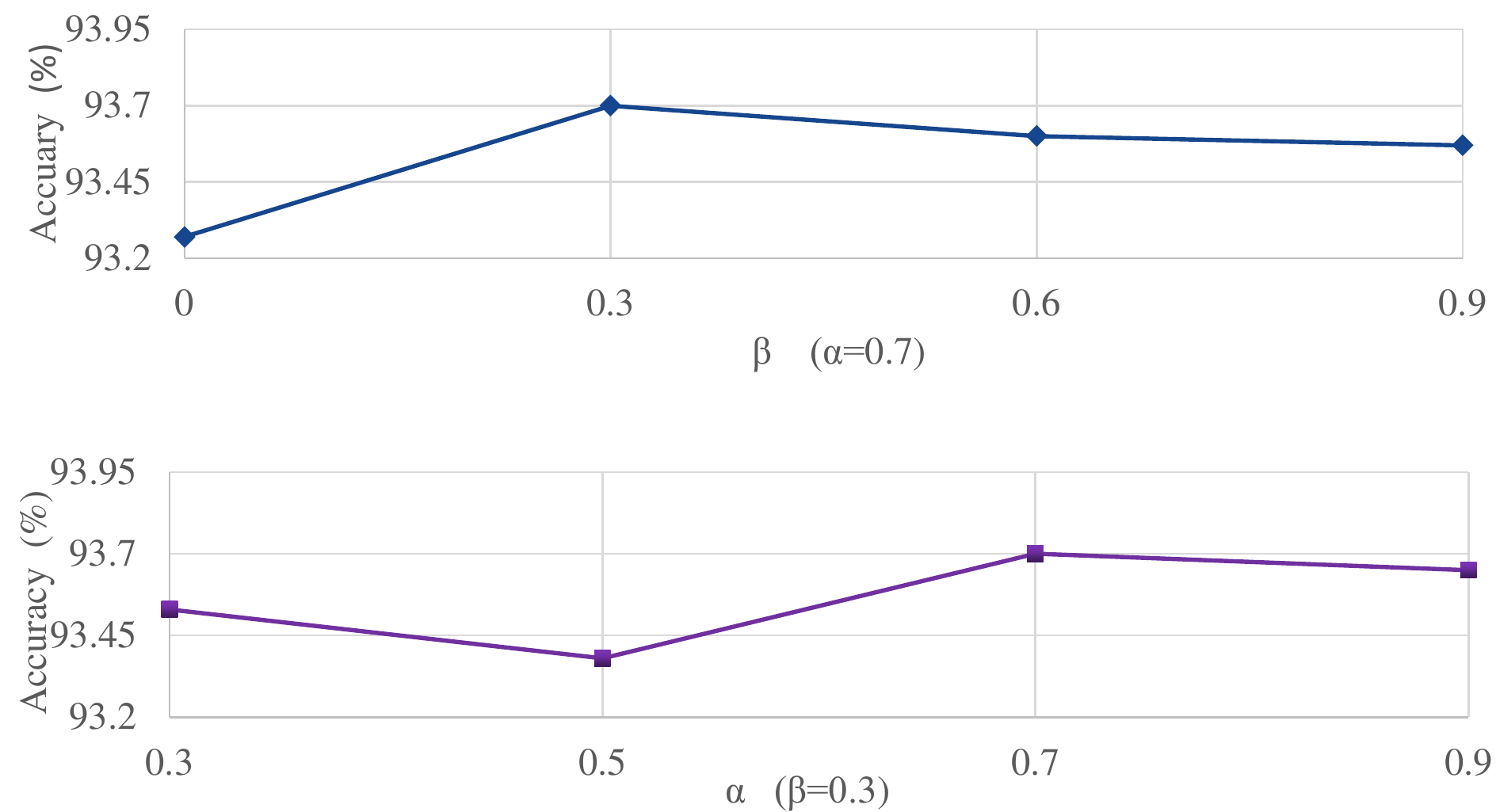}
\caption{The curves of performance are obtained by varying one hyper-parameter
%hyper parameter 
with the other parameter fixed to be the default value. The best performance is achieved when $\alpha = 0.7$, and $\beta = 0.3$.}
\label{fig_loss_alph}
\end{figure}

We conduct the diagnostic study to discuss the sensitiveness of the hyper-parameters ${\alpha}$ and ${\beta}$ in Eq.~\eqref{eq18}. Here, to enhance the efficiency of the diagnostic study, we use ResNeXt18 instead of ResNeXt101 as the RGB backbone and set $T=16$ without loss of generality. Specifically, we empirically tune them by ${\alpha}\in\{0.3, 0.5,0.7, 0.9\}$ and ${\beta}\in \{0, 0.3, 0.6, 0.9\}$, respectively. The corresponding results are shown in Figure~\ref{fig_loss_alph}. It can be found that the best performance of ESE-FN is achieved when $\alpha = 0.7$, and $\beta = 0.3$. Thus, we set $\alpha = 0.7$, and $\beta = 0.3$ in experiments. Moreover, when ${\beta}$=0, the multi-modal loss ${\mathcal L}$ is degraded to a basic action recognition loss ${{\mathcal L}}_{rs}$, where the performance is degraded significantly. This illustrates that the designed Multi-modal Loss is more effective than the basic loss.

%{\color{blue}To verify that our loss function can preserve the consistency of fusion modal ${loss}$ with single modal ${loss_r}$ and ${loss_s}$, we have curved the ${loss_r}$, ${loss_s}$ and ${loss}$.

%From Figure~\ref{fig_loss_cmp}, we can find that trend of fusion modal $loss$ is similar to that of the minimum of single modal $loss_r$. That shows our ML can preserve the consistency of fusion modal ${loss}$ with single modal ${loss_r}$ and ${loss_s}$.

%\begin{figure}[!t]
%\centering
%\includegraphics[scale=0.40]{IEEEtran/image/loss_cmp.pdf}
%\caption{The curves of single modal and fusion modal loss.}
%\label{fig_loss_cmp}
%\end{figure}

\subsection{Ablation Study}
\label{ab}
As mentioned before, Modal-fusion Net (M-Net), Channel-fusion Net (C-Net), and Multi-modal Loss (ML) are new components in the framework of the proposed ESE-FN.
Thus we conduct ablation study to evaluate the
superiority of M-Net, C-Net, and ML. We firstly set seven baselines, as follows,
	\begin{itemize}
		\item {\bf B1: Single modal for RGB} employs RGB backbone to extract the RGB features, which are directly used to train the softmax classifier. This aims to test the basic recognition performance by using the single-modal features in RGB videos.
		\item {\bf B2: Single modal for skeleton} employs skeleton backbone to extract the skeleton features, which are directly used to train the softmax classifier. This aims to test the basic recognition performance by using the single-modal features in skeleton sequences.
		\item {\bf B3: Multi modal for RGB and skeleton} employs RGB backbone and skeleton backbone to extract the RGB features and skeleton features, which are concatenated and fed into the softmax classifier. This aims to test the basic recognition performance by simply combining multi-modal features.
		\item {\bf B4: ESE-FN w/o C-Net} is a degraded version of ESE-FN by dropping out the channel-fusion module. This aims to show the superiority of M-Net.
		\item {\bf B5: ESE-FN w/o M-Net} is a degraded version of ESE-FN by dropping out the modal-fusion module. This aims to show the superiority of C-Net.
		\item {\bf B6: ESE-FN w/o ML} is a degraded version of ESE-FN by using the basic loss ${\mathcal L}_{rs}$ instead of ML ${\mathcal L}$, namely setting $\alpha=1$, and $\beta=0$. This aims to show the superiority of ML. 
		\item {\bf B7: the proposed ESE-FN}. 
	\end{itemize}

\begin{table}[!t]
\caption{{Ablation study for ESE-FN with different components.}}
\label{tab_abs}
\centering
\begin{tabular}{l|c|c|c|c|c|c}
\hline \hline
Baseline &   RGB          & Skeleton           & M-Net           & C-Net           & ML           & Accuracy (\%) \\ \hline
B1       &\checkmark      &                    &                 &                 &              & 93.5          \\ \hline
B2       &                &\checkmark          &                 &                 &              & 88.6         \\ \hline
B3       &\checkmark      &\checkmark          &                 &                 &              & 94.0        \\ \hline
B4       &\checkmark      &\checkmark          &\checkmark       &                 &\checkmark    & 95.3        \\ \hline
B5       &\checkmark      &\checkmark          &                 &\checkmark       &\checkmark    & 94.5        \\ \hline
B6       &\checkmark      & \checkmark         &\checkmark       &\checkmark       &              & 95.7            \\ \hline
B7       &\checkmark      & \checkmark         &\checkmark       &\checkmark       &\checkmark    & {\bf 95.9}      \\ \hline \hline
\end{tabular}
\end{table}

\begin{table}[!t]
\caption{Ablation study for comparing the performance of SENet (without Expansion) and ESE-FN.}
\label{tab2}
\centering
\begin{tabular}{l|c|c|c|c}
\hline \hline
Baseline &   Expansion   & Modal fusion     & Channel fusion    & Accuracy (\%) \\ \hline
A1       &               &                       & \checkmark             & 94.3           \\ \hline
A2       &               &\checkmark             &                        & 94.9         \\ \hline
A3       &               & \checkmark             &\checkmark              & 95.3        \\ \hline
A4       &\checkmark     &                       &\checkmark              & 94.5       \\ \hline 
A5       &\checkmark     &\checkmark             &                       & 95.3       \\ \hline
A6       &\checkmark     &\checkmark             &\checkmark               & 95.9        \\ \hline \hline
\end{tabular}
\end{table}

The comparison results among all baselines are shown in Table~\ref{tab_abs}. Compared with single-modal and multi-modal baselines, B3-B7 (using multi-modal features) outperform B1-B2 (using single-modal features), which validates that the multi-modal features contain more complementary information than the single-modal features. In particular, B4-B7 are equipped with the newly designed components, and perform better than B3 (simply combining multi-modal features). This validates that each component, i.e., M-Net, C-Net, or ML, is superior to help improving 
the representation ability of multi-modal feature fusion. Here, B6 integrated with M-Net and C-Net further improves by 0.4\%(1.2\%) over B4(B5) only with M-Net(C-Net). Obviously, ESE-FN (i.e., B7) integrated with all new components achieves the best performance. It is further validated that M-Net, C-Net, and ML can jointly improve the representation ability of multi-modal feature fusion.

%tab_mc-sfn_etri
\begin{table*}[!t]
\centering
\caption{{Comparison with current methods on ETRI-Activity3D dataset. Evolution Pose Map~\cite{liu2018recognizing} and FSA-CNN~\cite{jang2020etri} do not adopt the backbone module.}}
\label{tab_mc-sfn_etri}
\begin{tabular}{l|c|c|c}
\hline \hline
Methods                                      & Modalities                          & Backbones                             & Accuracy(\%) \\ \hline
Deep Bilinear Learning~\cite{hu2018deep}     & RGB+Depth+Skeleton                  & VGG16+RNN                             & 88.4             \\ \hline
Evolution Pose Map~\cite{liu2018recognizing} & RGB+Skeleton                        & -                                     & 93.6             \\ \hline
c-ConvNet~\cite{wang2018cooperative}         & RGB+Depth                           & VGG16                                 & 91.3             \\ \hline
FSA-CNN~\cite{jang2020etri}                  & RGB+Skeleton                        & -                                     & 93.7             \\ \hline 
\multirow{2}{*}{\textbf {ESE-FN~(Ours)}}  & \multirow{2}{*}{RGB+Skeleton}       & VGG16+Shift-GCN                         & 93.4             \\ \cline{3-4} 
                        &                                     & ResNeXt101+Shift-GCN                                       &\textbf{95.9}     \\ \hline \hline
\end{tabular}
\end{table*}

SENet is the most related work, where we have added the Expansion step based on SENet and pushed it into the multi-modal fusion task. Thus, we also conduct the ablation experiment to compare with the proposed ESE-FN and SENet by setting six baselines, as follows:
\begin{itemize}
        \item {\bf A1: SENet only for channel-wise fusion} uses SENet instead of C-Net for multi-modal feature fusion only in the channel-wise way. This can be seen as simply using Squeeze-Excitation for channel-wise fusion.
        \item {\bf A2: SENet only for modal-wise fusion} uses SENet instead of M-Net for multi-modal feature fusion only in the modal-wise way. This can be seen as simply using Squeeze-Excitation for modal-wise fusion.
		\item {\bf A3: SENet for modal and channel-wise fusion} uses SENet instead of M-Net and C-Net for multi-modal feature fusion in both the modal-wise and channel-wise ways. This can be seen as simply using Squeeze-Excitation for both modal-wise and channel-wise fusions.
		\item {\bf A4: ESE-FN only for channel-wise fusion} only uses C-Net for multi-modal feature fusion in channel-wise way.
		\item {\bf A5: ESE-FN only for modal-wise fusion} only uses M-Net for multi-modal feature fusion in modal-wise way.
		\item {\bf A6: ESE-FN (Ours)} uses M-Net and C-Net for multi-modal feature fusion in both the modal-wise and channel-wise ways.
	\end{itemize}

The comparison result of the ablation study is listed in Table~\ref{tab2}. We can find that 1) ESE-FN improves by 0.6\% over A3 (SENet for modal and channel-wise fusion), which illustrate that the Expansion step is effective; and 2) A3 outperforms both A1 and A2, as well as A6 outperforms both A4 and A5, which illustrate that our modal and channel-wise fusion strategy is effective and flexible for the SENet family.

\subsection{Results and Analysis}
We evaluate the performance of the proposed ESE-FN on the elderly activity recognition task by comparing it with the currently representative multi-modal methods, including Deep Bilinear Learning~\cite{hu2018deep}, Evolution Pose Map~\cite{liu2018recognizing}, c-ConvNet~\cite{wang2018cooperative}, and FSA-CNN~\cite{jang2020etri}. Specifically, Deep Bilinear Learning uses RGB frames, depth maps, and skeleton sequences; Evolution Pose Map uses 3D and 2D poses extracted from RGB frames and HeatMaps, respectively; c-ConvNet uses RGB frames and depth maps; FSA-CNN uses 2D and 3D skeletons extracted from RGB frames and depth maps. The comparison results of all methods are shown in Table~\ref{tab_mc-sfn_etri}, where some results have been reported in~\cite{jang2020etri}.
The proposed ESE-FN with the accuracy of 95.9\% achieves the state-of-the-art performance, which improves by 2.2\% (a relative
2.3\% improvement) over the previous SOTA method, namely FSA-CNN with the accuracy of 93.7\%. For a fair comparison, ESE-FN adopts the same backbone as the RGB feature extractor, and also performs better than the Deep Bilinear Learning method and c-ConvNet method, which illustrates the robustness of ESE-FN.

\begin{figure*}[!t]
\centering
\includegraphics[scale=0.55]{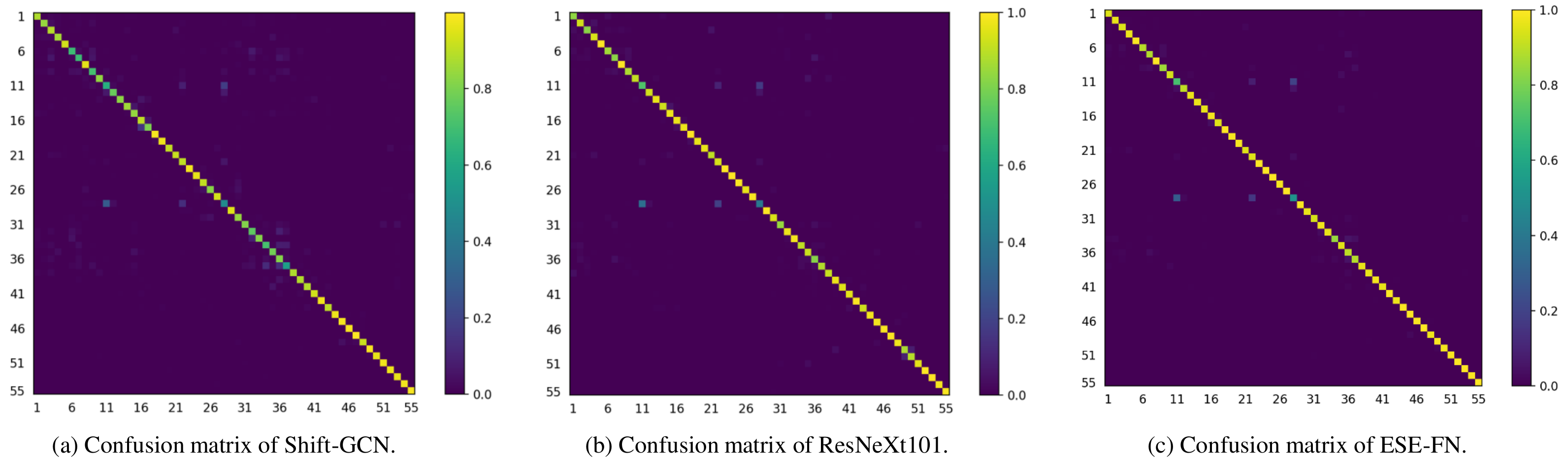}
\caption{{The confusion matrices obtained by Shift-GCN, ResNeXt101, and ESE-FN on the ETRI-Activity3D dataset. It should be noted that ETRI-Activity3D dataset does not provides the names of all class.}}
\label{fig_confusion_matrix_etri}
\end{figure*}

%The confusion matrix demonstrate the ESE-FN performance on ETRI-Activity3D dataset
\begin{figure*}[!t]
\centering
\includegraphics[scale=0.42]{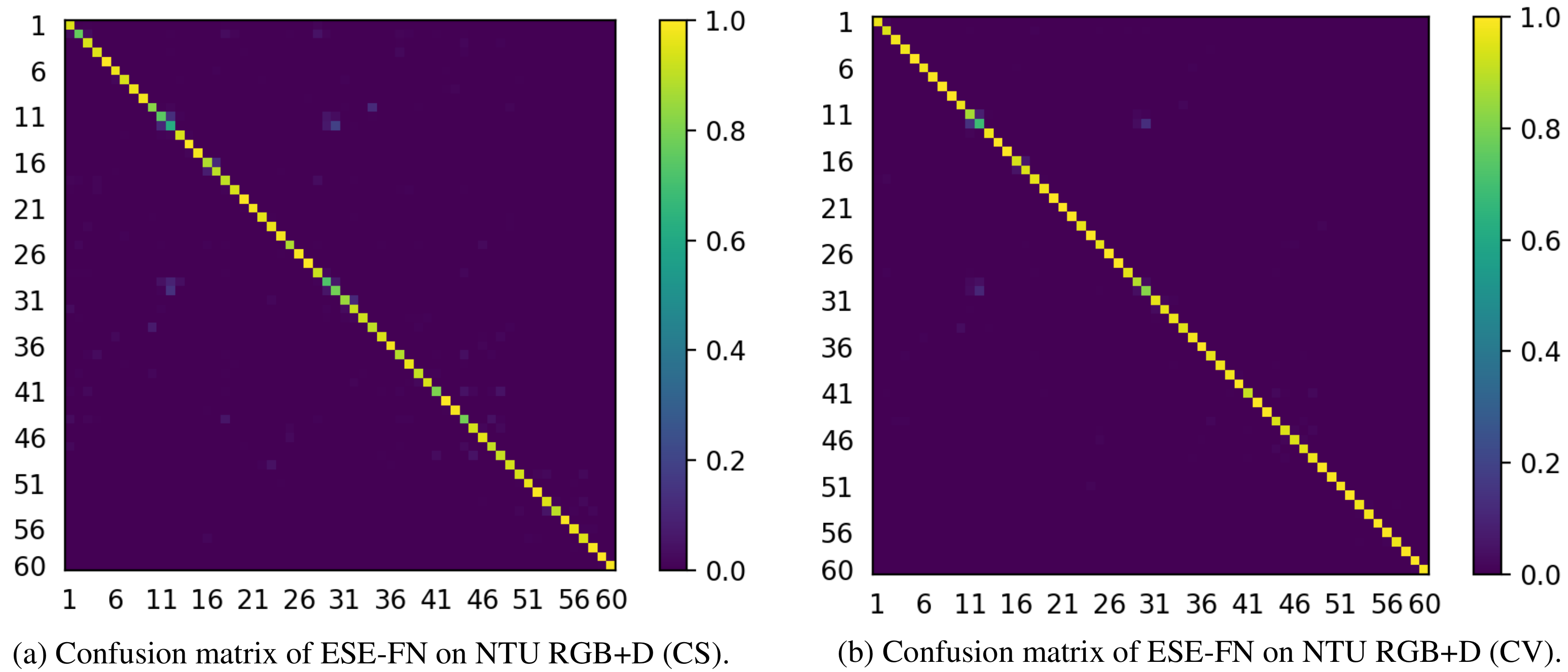}
\caption{{The confusion matrices obtained by ESE-FN on the NTU RGB+D dataset.}}
\label{fig_confusion_matrix_ntu}
\end{figure*}

In addition, the confusion matrices obtained by ESE-FN, Shift-GCN and ResNeXt101 are shown in Figure~\ref{fig_confusion_matrix_etri}. %This confusion matrix
First, the confusion matrix of ESE-FN shows that the color of the main diagonal is lighter than other spaces, which illustrates all classes of elderly activities can be well recognized by ESE-FN. Second, compared with the confusion matrices of ESE-FN, Shift-GCN, and ResNeXt101, we can find that the color of the main diagonal in Figure~\ref{fig_confusion_matrix_etri}(c) is lighter than the color of the main diagonal in Figure~\ref{fig_confusion_matrix_etri}(a) and (b). This shows that ESE-FN is more effective for recognizing the confusing elderly activities by capturing more discriminative multi-modal information.
%ESE-FN的混淆矩阵与RGB和SKeleton对比，阐述我们的ESE-FN确实可以区分老年人细微的动作，证明上面动作识别的例子。

%ESE-FN performance base on NUT RGB+D dataset compare with different method
\begin{table}[!t]
\centering
\caption{{Comparison with current methods on the NUT RGB+D dataset.}}
\label{tab_mc-sfn_ntu}
\begin{tabular}{l|c|c}
\hline \hline
Method                                 & CS (\%) & CV (\%) \\ \hline
IndRNN {\cite{li2018independently}}                   & 81.8   & 88.0   \\ \hline
Beyond Joint {\cite{wang2018beyond}}                  & 79.5   & 87.6   \\ \hline
SK-CNN {\cite{li2017skeleton}}                        & 83.2   & 89.3   \\ \hline
ST-GCN {\cite{yan2018spatial}}                        & 81.5   & 88.3   \\\hline
Motif ST-GCN {\cite{wen2019graph}}                    & 84.2   & 90.2   \\\hline
Ensem-NN {\cite{xu2018ensemble}}                      & 85.1   & 91.3   \\\hline
MANs {\cite{li2021memory}}                            & 83.0   & 90.7   \\\hline
HCN {\cite{li2018co}}                                 & 86.5   & 91.1   \\\hline
Deep Bilinear Learning {\cite{hu2018deep}}            & 85.4   & -      \\\hline
Evolution Pose Map {\cite{liu2018recognizing}}        & 91.7   & -      \\\hline
c-ConvNet {\cite{wang2018cooperative}}                & 82.6   & -      \\\hline
FSA-CNN {\cite{jang2020etri}}                         & 91.5   & -      \\\hline
STGR-GCN {\cite{li2019spatio}}                        & 86.9   & 92.3   \\\hline
2s-AGCN {\cite{shi2019two}}                           & 88.5   & 95.1   \\\hline
DGNN {\cite{shi2019skeleton}}                         & 89.9   & 96.1   \\\hline
Shift-GCN {\cite{cheng2020skeleton}}                  & 90.7   & 96.5   \\\hline
MS-G3D Net {\cite{liu2020disentangling}}              & 91.5   & 96.2   \\\hline
\textbf{ESE-FN~(Ours)}                              & \textbf{92.4}    &\textbf{96.7}      \\ \hline \hline
\end{tabular}
\end{table}

\subsection{Extended Experiment on Normal Action Recognition}
\label{r-n}
To test the generalization of the proposed ESE-FN, we also conduct comparative experiments between the proposed ESE-FN and the other advanced methods in terms of the normal action recognition task. We use the NTU RGB+D dataset~\cite{shahroudy2016ntu} as the benchmark. The NTU RGB+D dataset is a large-scale dataset collected by three Kinect cameras from different views concurrently, and has been widely used for evaluating RGB-based or skeleton-based action recognition methods. It includes 56,880 skeleton sequences and 60 different classes from 40 distinct subjects. NTU RGB+D dataset provides two standard evaluation protocols, i.e., Cross-Subject (CS) setting, and Cross-View (CV) setting. In the CS setting, the training set includes 40,320 samples performed by 20 subjects, and the testing set includes 16,560 samples performed by the other 20 subjects. In the CV setting, the training set includes 37,920 samples obtained from camera views 2 and 3, and the testing set includes 18,960 samples obtained from camera view 1. 

Table~\ref{tab_mc-sfn_ntu} shows the recognition accuracies obtained by different methods on the NTU RGB+D dataset, where the performance of the other methods have been reported in~\cite{jang2020etri}. We can see that ESE-FN is comparable to the alternatives, including RNN-based methods (e.g., IndRNN~\cite{li2018independently}, Beyond Joint~\cite{wang2018beyond}), CNN-based methods (e.g., SK-CNN~\cite{li2017skeleton}, MANs~\cite{li2021memory}, Deep Bilinear Learning~\cite{hu2018deep}), and GCN-based methods (e.g., ST-GCN~\cite{yan2018spatial}, Shift-GCN~\cite{cheng2020skeleton}). Specifically, ESE-FN improves by 0.7\% and 0.2\% over the SOTA methods, namely Evolution Pose Map with the accuracy of 91.7\% and Shift-GCN with the accuracy of 96.5\% on CS and CV settings, respectively. In addition, we also show the confusion matrices obtained by ESE-FN on NTU RGB+D (CS and CV) in Figure~\ref{fig_confusion_matrix_ntu}(a) and (b). %This confusion matrices shows
These confusion matrices show that the color of the main diagonal is lighter than other spaces, which illustrates all classes of normal actions can be well recognized by ESE-FN. It can be concluded that the proposed ESE-FN is generalized to recognize normal actions.

\section{Conclusion}
\label{C}
 In this work, we propose a novel Expansion-Squeeze-Excitation Fusion Network (ESE-FN) to well address the problem of elderly activity recognition by learning modal and channel-wise Expansion-Squeeze-Excitation (ESE) attentions for attentively fusing the multi-modal features in the modal and channel-wise ways, respectively. Overall, the framework of ESE-FN consisting of the feature extractor module, modal-fusion module, channel-wise module, and multi-modal loss can be summarised as two main insights. First, Modal-fusion Net (M-Net) and Channel-fusion Net (C-Net) can capture the modal and channel-wise dependencies among features for enhancing the discriminative ability of features via Modal and channel-wise ESEs. Second, Multi-modal Loss (ML) can enforce the consistency between the single-modal features and the fused multi-modal features by additionally bringing in the penalty of difference between the minimum prediction losses on the single modalities and the prediction loss on the fused modality. Experimental results on %both of
both elderly activity recognition and normal action recognition tasks demonstrate that the proposed ESE-FN achieves the SOTA performance compared with the other competitive methods. To the best of our knowledge, ESE-FN is the first work that adopts the combination of expanding and squeezing to fully interacting features from the local and global views for learning nonlinear attention. 

{\bf Discussions.} We refer to SENet's idea that enhances the feature quality of different channels via nonlinear attention. SENet only uses a single global pooling operation to get a representation of channel information, which can be seen as a down-sampling interaction of features. In this work, we consider both the up-sampling and down-sampling interaction of features and present a new Expansion step, which can be further used to expand the modal information and channel information. The idea of up-sampling and down-sampling via Expansion and Squeeze is the first attempt in the SENet family, which will be insightful for the other researchers. Moreover, as one member of the SENet family, ESE-FN is not limited only to the task of conventional feature learning and has been successfully extended for multi-modal feature fusion learning. This can be seen as an exemplary case that shows SENet for multi-modal feature fusion learning.

\section{Acknowledgements}
The work is supported by the National Key RI\&D Program of China (No. 2018AAA0102001), the Natural Science Foundation of Jiangsu Province (Grant No. BK20211520), and the National Natural Science Foundation of China (Grant No. 62072245, and 61932020).

%explore modal-wise fusion and channel-wise fusion of multi-modal features by learning modal and channel-wise nonlinear attention for capturing the modal and channel-wise dependencies among features. %In future, we attempt to  evolve the proposed method explore into more flexible and full fusion version for fusing multi-modal features.

% or


%\appendices

%\section{}

%\section*{Acknowledgment}


\ifCLASSOPTIONcaptionsoff
  \newpage
\fi



%\bibliographystyle{IEEEtran}
\bibliographystyle{IEEEtran}
\bibliography{IEEEabrv,ref}

\end{document}